\pdfoutput=1

\documentclass[11pt]{article}

\usepackage[]{acl} 

\usepackage{times}
\usepackage{latexsym}

\usepackage{amsmath}
\usepackage{graphicx}
\usepackage{subcaption}
\usepackage{booktabs}
\usepackage{longtable}
\usepackage{rotating} 
\usepackage[section]{placeins} 
\newcommand{\phenomenon}[1]{{\fontfamily{cmtt}\selectfont #1}}
\usepackage{multirow}
\usepackage{linguex} 
\usepackage{enumitem}
\usepackage{tablefootnote}
\usepackage{float}
\newcommand\tablerotation{20}

\usepackage[T1]{fontenc}

\usepackage[utf8]{inputenc}

\usepackage{microtype}

\usepackage{inconsolata}

\title{Interpretability of Language Models via Task Spaces}

\author{Lucas Weber* \\
  University Pompeu Fabra \\
  \texttt{lucasweber000@gmail.com} \\\And
  Jaap Jumelet \\
  ILLC, University of Amsterdam\\
  \texttt{jumeletjaap@gmail.com} \\\AND
  Elia Bruni$^{\dagger}$ \\
  Osnabrück University \\
  \texttt{elia.bruni@gmail.com} \\\And
  Dieuwke Hupkes$^{\dagger}$ \\
  Meta \\
  \texttt{dieuwkehupkes@meta.com}
}

\begin{document}
\maketitle
\begingroup\def\thefootnote{*}\footnotetext{Now affiliated with \textbf{Fraunhofer IIS}; corresponding author}\endgroup
\begingroup\def\thefootnote{$\dagger$}\footnotetext{Shared senior authorship}\endgroup
\begin{abstract}
The usual way to interpret language models (LMs) is to test their performance on different benchmarks and subsequently infer their internal processes.
In this paper, we present an alternative approach, concentrating on the \textit{quality} of LM processing, with a focus on their language abilities.
To this end, we construct `linguistic task spaces' -- representations of an LM's language conceptualisation -- that shed light on the connections LMs draw between language phenomena.
Task spaces are based on the interactions of the learning signals from different linguistic phenomena, which we assess via a method we call `similarity probing'.
To disentangle the learning signals of linguistic phenomena, we further introduce a method called `fine-tuning via gradient differentials' (FTGD).
We apply our methods to language models of three different scales and find that larger models generalise better to overarching general concepts for linguistic tasks, making better use of their shared structure. 
Further, the distributedness of linguistic processing increases with pre-training through increased parameter sharing between related linguistic tasks. 
The overall generalisation patterns are mostly stable throughout training and not marked by incisive stages, potentially explaining the lack of successful curriculum strategies for LMs.
\end{abstract}

\section{Introduction}
\label{ch2:introduction}

Recently, language models (LMs) have reached a level of sophistication in language production where their output is often indistinguishable from human-generated language \citep[][]{liang2022holistic}. 
However, the complexity inherent in language production means that effective models are also inherently complex, making them challenging to interpret.

\begin{figure}[h!]
\centering
\includegraphics[width=\linewidth]{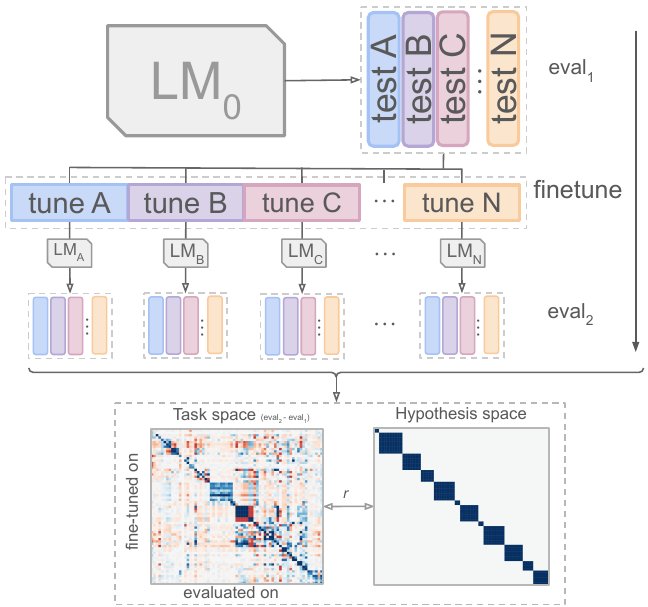}
\caption{The process of similarity probing to obtain a task space based on transfers: 1. Evaluate the untuned LM on all tasks (eval\textsubscript{1}); 2. Tune one LM for each task; 3. Re-evaluate the LMs on all tasks (eval\textsubscript{2}). Calculate all transfers (eval\textsubscript{2} - eval\textsubscript{1}) and compare the resulting transfer task space to a hypothesized set of transfers (Hypothesis space).}
\label{fig:transfer_spaces}
\end{figure}

Commonly, linguistic interpretability involves assessing an LM's ability through simple evaluation tasks like grammatical acceptability judgments of various language constructions \citep[e.g.][]{linzen2016assessing, Marvin2018TargetedModels}. 
While these methods inform us about a model's \textit{performance}, they do not provide insights into the \textit{quality} of the model's solutions.
This is especially the case when error analysis is not possible due to high model performance.
However, it is the quality of processing that is interesting from the viewpoint of the interpretability researcher, the cognitive scientist or linguist.
Here, we introduce a method to interpret the language processing of LMs holistically. We show how linguistic knowledge in LMs interconnects.
We build upon the framework of \citet{weber-etal-2021-language} that proposes to consider linguistic phenomena as `tasks' an LM has to optimise and allows us to analyse the interactions of those tasks, similar to ideas from multi-task learning (MTL). 
For example, consider the following sentences:

\ex. John did \textit{not} see \textbf{anything}.

\ex. \textit{If} John sees \textbf{anything}, he will be surprised.

In both sentences, a downward-entailing environment (negation vs. conditional) allows for the negative polarity item (NPI) \textit{anything} to be used.
Understanding whether it is acceptable to produce an NPI in either sentence can be considered a different task. 
LMs can use different rules to solve these tasks:
an LM might learn the co-occurrence statistics of certain trigger words (e.g. `\textit{not}' vs. `\textit{if}') with NPIs. On the other hand, it might generalise to a more abstract linguistical conceptualisation and understand that both -- negation and conditionals -- create downward-entailing environments permitting NPIs. 
With either rule, the model resolves acceptability judgements correctly, while the \textit{quality} of both solutions is decisively different.
Hence, assessing the generalisation of linguistic tasks reveals how LMs conceptualise language.

Similar to `task spaces' in MTL (more details in §~\ref{ch2:background:sim_spaces}), we can represent an LM's generalisation behaviour in a \textit{linguistic task space}, a multi-dimensional space relating linguistic tasks according to their similarity.
To construct linguistic task spaces, we introduce \textit{similarity probing}, a method to estimate linguistic similarity. This method involves selectively fine-tuning LMs on specific linguistic tasks and assessing the impact of the fine-tuning on other tasks (see Figure~\ref{fig:transfer_spaces}), as well as the alignment of tasks in gradient space.
We extricate single linguistic tasks from their entanglement in natural language via a method we call \textit{fine-tuning via gradient differentials (FTGD)}.
FTGD selectively updates a small, relevant subspace of parameters with `gradient differentials'.

The contributions of this paper can be summarised as follows:
\begin{enumerate}[itemsep=-.15em]
    \item Propose linguistic task spaces as an interpretability method for deeper model understanding and as a tool for linguistic theory testing.
    \item Introduce \textit{FTGD}, a technique to disentangle linguistic tasks from their language context and selectively fine-tune them in LMs.
    \item Introduce \textit{similarity probing}, an efficient method for generating large linguistic task spaces.
    \item Analyze the development of language conceptualisation of LMs throughout pre-training by constructing language task spaces at various stages of LM pre-training.
\end{enumerate}

\section{Background and related work}
\label{ch2:background}
In this section, we summarise related work and additional background on task spaces in MTL (§~\ref{ch2:background:sim_spaces}),  the use of LMs for linguistic theorizing (§~\ref{ch2:background:lang_spaces}) and fine-tuning machine learning models in low-dimensional subspaces (§~\ref{ch2:background:tuning_in_ld_space}).
\subsection{Task-similarity spaces in MTL}
\label{ch2:background:sim_spaces}
In MTL, the transfer between different tasks is thought to be determined by their `similarity' \citep{ben2008notion}.
Constructing similarity spaces and task taxonomies to determine which tasks should be trained together has been a prominent goal in the literature.
One of the earliest examples of constructing task-similarity spaces can be found in \citet{Thrun1996DiscoveringAlgorithm}.
More recently, \citet{Zamir_2018_CVPR, standley2020tasks} constructed task taxonomies for computer-vision tasks based on the transferability of task-specific representations. 
Similarly, \citet{achille2019task2vec} create `task embeddings' for visual classification tasks by comparing their task structure through Fisher Information Matrices.
From a theoretical perspective, \citet{lee2021continual} investigate task similarity using synthetic tasks in a controlled setting, finding that their similarity measure can predict learning outcomes.

\subsection{Linguistic spaces}
\label{ch2:background:lang_spaces}
Their ability to consistently construct acceptable language made LMs interesting, explicit linguistic theories \citep{baroni2022proper}.
However, similar to humans \citep{watson1913psychology,titchener1912schema,nisbett1977telling}, LMs cannot introspectively report their internal processes.
Consequently, there has been growing interest in developing methods to gain theoretical insights by analysing internal processes of LMs in what has been described as `synthetic linguistics' \citep{chowdhury-zamparelli-2019-lstm}.
A collection of interpretability work assumes implicit linguistic similarity spaces in LMs revealed in their generalisation behaviour:
\citet{weber-etal-2021-language} demonstrate how language models generalize across linguistically related constructions, suggesting an implicit task hierarchy within broad tasks like language modeling.

\citet{chowdhury-zamparelli-2019-lstm} observe that pre-trained models more easily learn grammatical than ungrammatical structures, showing how LMs generalise across consistent linguistic structure.
\citet{prasad2019using} and \citet{sinclair2022structural} use priming experiments to determine the relationship between different linguistic tasks and recover their hierarchical organization.
\citet{perez2021evolution} fine-tune LMs on various downstream tasks and evaluate the effects on their syntactic understanding.
\citet{muller2023subspace} probe for linguistic subspaces in language models using information theoretic probes.
To our knowledge, we conduct the first attempt to \textit{explicitly} construct a linguistic similarity space.
Ultimately, our linguistic spaces are akin to knowledge representations in conceptual spaces \citep{gardenfors2004conceptual,gardenfors2014geometry}, popular in the cognitive sciences.

\subsection{Fine-tuning in low-dimensional subspaces}
\label{ch2:background:tuning_in_ld_space}

Recently, the idea that tasks can be fine-tuned in low-dimensional subspaces of overparameterised models has gained popularity.
A range of previous work shows how tasks can be effectively trained using projections into low-dimensional subspaces \citep{li2018measuring, aghajanyan2020intrinsic, gressmann2020improving, hu2021lora, li2022low, zhang-etal-2023-fine}.
Other research demonstrates the reducibility of tasks to small subnetworks by learning discrete maskings for task-irrelevant parameters or activations \citep{frankle2018lottery, mallya2018piggyback, sanh2020movement, zhang2021disentangling, zhao2020masking, csordas2020neural, guo2020parameter,  chintam2023identifying}, without explicitly projecting representations into lower-dimensional space.


\section{Methods}
\label{ch2:methods}
In this paper, we connect work on task spaces in MTL \citep{Zamir_2018_CVPR,standley2020tasks,achille2019task2vec} with the idea of linguistic similarity spaces \citep{weber-etal-2021-language} in a method we call \textbf{similarity probing}.
Similarity probing consists of three steps:
First, we evaluate our untuned model on a wide range of linguistic phenomena (i.e. a `linguistic task').
Then, we selectively tune a separate LM on each linguistic task.
Finally, we evaluate each model again on all linguistic tasks and assess the tuning's impact in terms of performance transfers and compare different properties of the gradient updates (more details follow in §~\ref{ch2:methods:sim_probing}). 
Fine-tuning a linguistic phenomenon is not straightforward.
For that reason, we start by introducing a modified training procedure we will call `fine-tuning via gradient differentials' (§~\ref{ch2:methods:contr_tuning}).

\subsection{Fine-tuning via gradient differentials (FTGD)}
\label{ch2:methods:contr_tuning}

The major problem when fine-tuning an LM on a linguistic phenomenon is what we here call `linguistic entanglement'. 
\paragraph{Linguistic entanglement}
Within language data, linguistic tasks are necessarily interwoven \citep{weber-etal-2021-language}.
For example, certain tasks are present in every sentence (e.g. \phenomenon{subject verb agreement [SVA]}).
This presents a challenge if we want to use natural language data to selectively fine-tune a separate task \phenomenon{A} since any potential data point to train task \phenomenon{A} necessarily also contains information on \phenomenon{SVA}.
The learning signals of \phenomenon{A} and of \phenomenon{SVA} are overlapping and can not be unambiguously attributed to either task.

On the other hand, natural language data is also rich in spurious correlations between different task distributions.  
For example, two tasks \phenomenon{A} and \phenomenon{B} might occur in similar contexts or frequently share vocabulary in their realisations.
The similarity in learning signal between \phenomenon{A} and \phenomenon{B} in such cases may solely be due to these spurious correlations instead of any conceptual similarity.


\begin{figure}[h!]
\centering
\includegraphics[width=\linewidth]{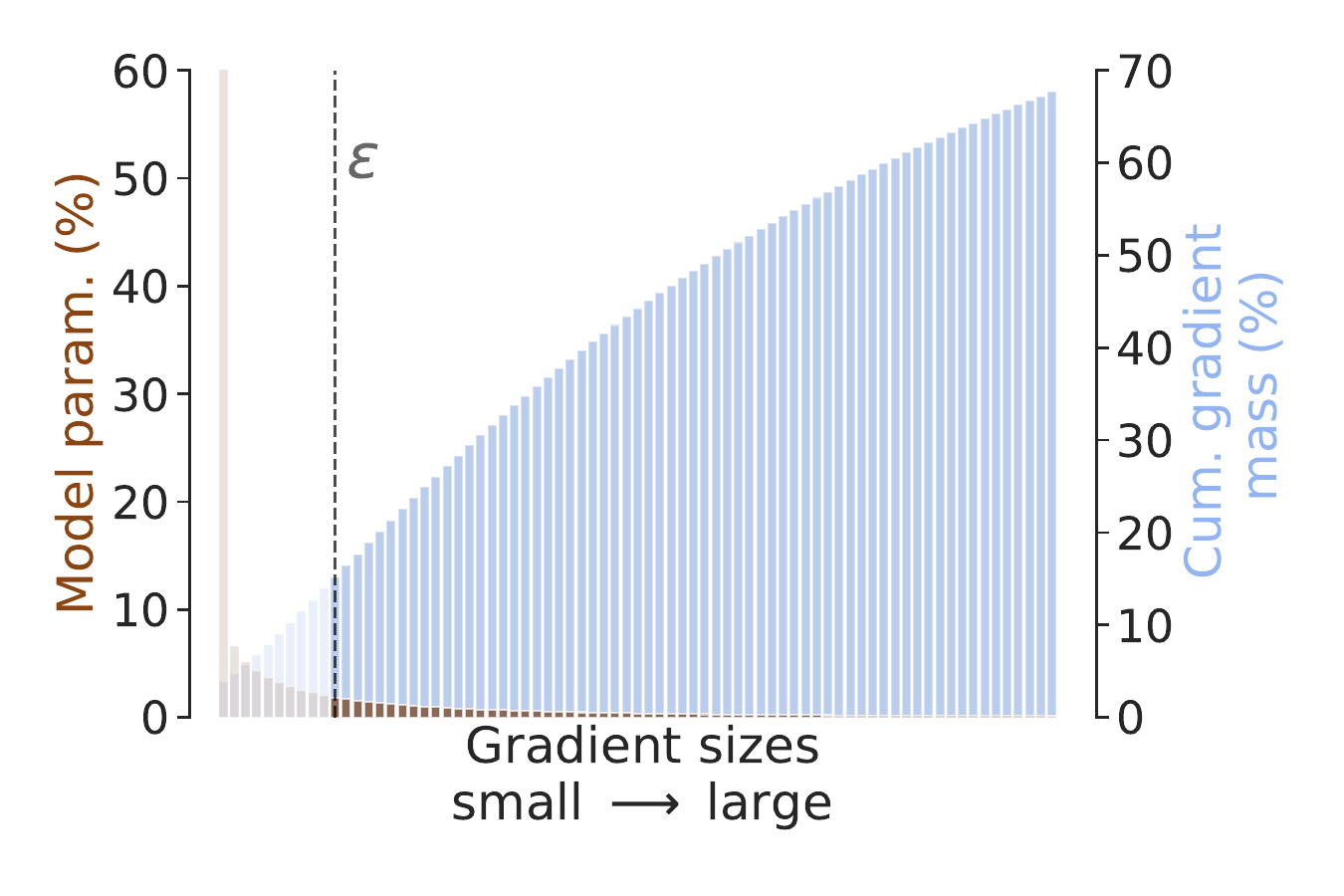}
\caption[]{Parameter bins ranging from small to large gradients. Only a small amount of parameters carry the largest portion of gradient mass. Our cut-off ($\epsilon$) maintains a large portion of gradient mass while reducing the amount of trained parameters significantly.}
\label{fig:parameter_count_vs_gradient_mass}
\end{figure}

\begin{figure*}[h!]
\centering
\begin{subfigure}[t]{0.42\linewidth} 
\includegraphics[width=\linewidth]{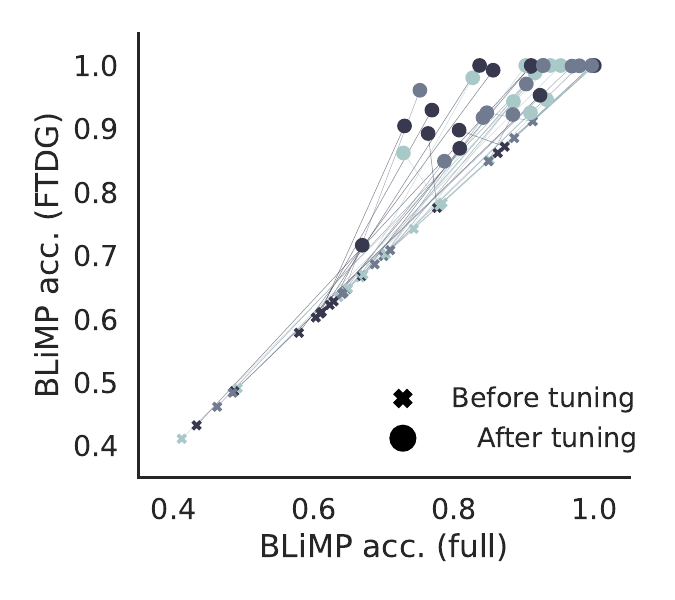}
\vspace{-.65cm}
\caption{}
\label{subfig:acc_contrastive_vs_full}
\end{subfigure}
\hspace{-1cm}
\hfill
\centering
\begin{subfigure}[t]{0.39\linewidth} 
\raisebox{6pt}{\includegraphics[width=\linewidth]{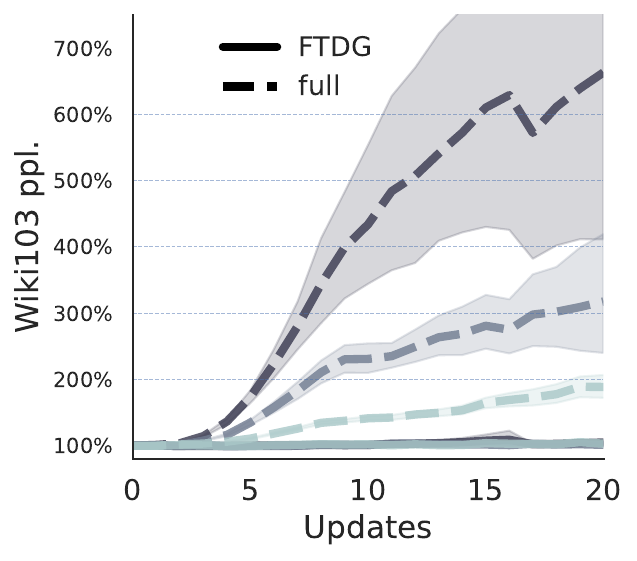}}
\vspace{-.65cm}
\caption{}
\label{subfig:wiki_val_contr_vs_full}
\end{subfigure}
\begin{subfigure}[t]{0.14\linewidth} 
\raisebox{70pt}{\includegraphics[width=\linewidth]{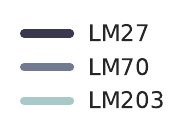}}
\end{subfigure}
\vspace{-.15cm}
\caption{(a) BLiMP accuracy per phenomenon before and after fine-tuning using full gradients or our gradient difference method. FTGD is either as effective or more effective in improving benchmark performance on all phenomena. (b) The relative increase in perplexity (ppl) on the \textit{wiki103} validation set during the fine-tuning process of models trained for 20 epochs. FTGD barely affects perplexity, while full gradients are highly disruptive.}
\label{fig:contrastive_tuning}
\vspace{-.25cm}
\end{figure*}

\paragraph{Disentangling linguistic tasks}
Our method builds on the assumption that gradients in LM training are a linear combination of an ensemble of `subgradients', representing different linguistic tasks.
Following this assumption, we can isolate a specific linguistic task by subtracting two gradients that only differ in that task's specific subgradient, effectively eliminating all other subgradients.
To obtain minimally different gradients, we make use of `minimal pair' sentences. 
A minimal pair consists of two almost identical sentences, only distinguished by a minimal difference that renders one of them ungrammatical with respect to a specific linguistic task.
An example of a minimal pair for NPI licensing through negation is

\ex. \hspace{1pt} John did \textit{not} see \textbf{anything}.

\ex. * \hspace{1pt} John did see \textbf{anything}.

where (4) differs minimally from (3) to render it unacceptable.
Using minimal pairs as training data, we proceed in the following way:
At every update, we calculate the gradients \( g^{+}(\theta) \) for the grammatical examples and \( g^{-}(\theta) \) for their ungrammatical counterparts, with \( \theta \) being our model parameters.
We then calculate the gradient differentials \( g^{\Delta} = g^{+} - g^{-} \).

Estimating similarity in the resulting high-dimensional gradient space is challenging  \citep[see e.g. ][]{beyer1999nearest,zimek2012survey}.
Therefore, we additionally reduce the gradients to a small parameter subspace by dropping parameters for which the differential \( g^{\Delta}(\theta) \) does not differ from 0 by a margin of \( \epsilon = 10^{-3} \): 

\[ \theta_0 = \{\theta : |g^{+}(\theta) - g^{-}(\theta)| > \epsilon\} \]

In other words, we select those parameters \( \theta_0 \) where the gradients for positive and negative examples are sufficiently different.
We then fine-tune the model by using only \( g^{\Delta}(\theta_0) \).
With this approach, we reduce the number of trainable parameters to an average of 5\% of the full parameters while maintaining 81\% of the gradient mass (see Figure~\ref{fig:parameter_count_vs_gradient_mass}).

\subsection{Similarity Probing}
\label{ch2:methods:sim_probing}
We determine the similarity between tasks \phenomenon{A} and \phenomenon{B} by examining various aspects of how LMs learn them.
Following the MTL literature, we concentrate on \textit{transfer learning} \citep{Zamir_2018_CVPR,standley2020tasks} and the \textit{gradient alignment} \citep{yu2020gradient} between \phenomenon{A} and \phenomenon{B}.

\paragraph{Transfer probing}
We determine the transfer between \phenomenon{A} and \phenomenon{B} by fine-tuning a language model on \phenomenon{A} and measuring the performance on a benchmark test of \phenomenon{B} before and after the fine-tuning (see Figure~\ref{fig:transfer_spaces}). 
Fine-tuning on task \phenomenon{A} may have three potential influences on task \phenomenon{B}. 
We interpret them as follows:
\begin{enumerate}[itemsep=-.15em]
    \item The performance of \phenomenon{B} increases: \phenomenon{A} and \phenomenon{B} are related and have high similarity.
    \item The performance of \phenomenon{B} decreases: \phenomenon{A} and \phenomenon{B} are related and have low similarity. 
    \item The performance of \phenomenon{B} is unchanged: \phenomenon{A} and \phenomenon{B} are unrelated. 
\end{enumerate}

\noindent We normalise\footnote{For negative transfers, we normalise by the \textit{maximally possible accuracy decrease} on the benchmark (i.e. the pre-fine-tuning accuracy), and for positive transfers, we normalise by the \textit{maximally possible accuracy increase} on the benchmark (i.e., 1 - the pre-fine-tuning accuracy).} all transfers to mitigate any floor and ceiling effects. 
Assessing similarity in this way has the advantage of being technically easier to apply and does not require a machine with a large memory.

\paragraph{Gradient Probing}
We can also directly relate tasks in the parameter space:
we compare the \textit{overlap} of different task subspaces and the \textit{alignment} of gradients \( g^{\Delta} \) within those subspaces.
Following \citet{yu2020gradient}, we predict:

\begin{enumerate}[itemsep=-.15em]
    \item \phenomenon{A} and \phenomenon{B} have great subspace overlap and $g^{\Delta}$ is aligned: \phenomenon{A} and \phenomenon{B} are similar and will benefit each other.
    \item \phenomenon{A} and \phenomenon{B} have a great subspace overlap, but $g^{\Delta}$ is not aligned: \phenomenon{A} and \phenomenon{B} are dissimilar and will interfere.
    \item \phenomenon{A} and \phenomenon{B} have a small or no subspace overlap: \phenomenon{A} and \phenomenon{B} are unrelated and will not interact.
\end{enumerate}

\noindent We determine the \textbf{overlap} between subspaces of tasks \phenomenon{A} and \phenomenon{B} by calculating their Jaccard-similarity $J_{sim}$
\[ J_{sim}(\theta_0^A, \theta_0^B) = \frac{|\theta_0^A \cap \theta_0^B|}{|\theta_0^A \cup \theta_0^B|} \]
and use cosine similarity (CS) to measure gradient alignment between tasks.
The gradient-based method enables us to get a more detailed insight into the interaction between tasks.

\section{Experiments}
\label{ch2:experiments}
In the empirical analysis of our method, we pre-train three different generative LMs up to various stages. 
Then, we test the FTGD on a trained-out checkpoint to ensure that it works as intended.
Subsequently, we apply it to all intermediate checkpoints to interpret the change in the LMs' language conceptualisation throughout the training process. 

\subsection{Training details}
\label{ch2:experiments:details}

\paragraph{Data}
\label{ch2:methods:data}
We pre-train our LMs on the standard split of a common English Wikipedia corpus \citep[;][]{merity2016pointer}.
For probing their linguistic ability, we use the BLiMP corpus \citep[][]{warstadt2020blimp}.
BLiMP consists of minimal pairs for 13 higher-level linguistic \emph{phenomena} in the English language, which can be subdivided into 67 distinct realisations called \emph{paradigms}.
Each paradigm contains 1000 individual minimal pairs, sizing the whole corpus at 67,000 data points.
For our experiments, we consider every paradigm as a separate linguistic task.
We use 85\% of each paradigm's data for probe training and retain 15\% for evaluation.

\paragraph{Models and pre-training}
We employ decoder-based generative transformer language models based on the fairseq library \citep{ott2019fairseq}.
We consider three models of different sizes with $\sim$27M, $\sim$70M  and $\sim$203M trainable parameters\footnote{Hyperparameters for ({\small\textsf{LM27}}, {\small\textsf{LM70}} and {\small\textsf{LM203}}):  layers = (3, 6, 12), hidden- and embedding-size = (256, 512, 1024), attention-heads = (4, 8, 16), ffn size = (1024, 2048, 4096)} respectively, with all other hyperparameters kept constant across models\footnote{Hyperparameters: batch size = 16, dropout = 0.1, learning rate = 0.0001; Optimiser: Adam \citep{kingma2014adam}}.
After 20 epochs of pre-training, we reach average final perplexities of 65.21, 38.32 and 27.61 on the \textit{wiki103} validation set of training across 5 runs.

\paragraph{FTGD}
We adapt the setup for fine-tuning in the probing phase to avoid potential confounds: 
we switch to plain stochastic gradient descent (SGD) to preempt interference of Adam's momentum terms with the probing process. 
Additionally, we change the batch size to 850 to contain the entire training data for the probed paradigm to minimize the influence of individual data point variations on the learning signal.
We fine-tune on the probed paradigm until the model's performance on the validation set converges\footnote{The stopping criterion is defined as current performance being lower or equal to the average of the last five steps.} or we reach a maximum of 20 updates.

\subsection{Results}
\label{ch2:experiments:results}

We first show how FTGD compares to full-gradient fine-tuning (§~\ref{ch2:experiments:results:contr_tuning}), continue with an analysis of the similarity probing method (§~\ref{ch2:experiments:results:trans_space}) and end with the analysis of the development of similarity space throughout the LM pertaining process (§~\ref{ch2:experiments:results:learning_process}). 

\subsubsection{FTGD}
\label{ch2:experiments:results:contr_tuning}
The desideratum of our method is that it improves a specific linguistic task in isolation. 
This requires effectiveness and selectivity: it improves the model's performance on a specific linguistic task while not interfering with unrelated capacities.
We assess effectiveness by comparing FTGD with regular fine-tuning using the full gradients $g^+$ and observe that the difference method achieves equivalent or higher fine-tuning performance (see Figure~\ref{subfig:acc_contrastive_vs_full}).
To assess selectivity, we compare how much both fine-tuning methods interfere with the LM's general ability to generate language.
We measure this ability via the LM's perplexity on the \textit{wiki103} validation set and find FTGD to be much less disruptive than full gradient fine-tuning (see Figure~\ref{subfig:wiki_val_contr_vs_full}).
Both results show how FTGD is indeed effective and selective.

\subsubsection{Linguistic task spaces} 
\label{ch2:experiments:results:trans_space}

After applying our similarity probing method, we obtain linguistic task spaces containing similarity values between all possible pairings of BLiMP paradigms. The heatmap in Figure~\ref{fig:transfer_spaces} visualises a transfer space for a {\small\textsf{LM203}} model. 
From here, we can take two perspectives on the analysis of linguistic task spaces: we can use them to test linguistic hypotheses or to interpret the linguistic conceptualisation of the LM.
While we will concentrate on the latter use for the remainder of the paper, we will touch upon the possibility of doing linguistic hypothesis testing in the discussion.

\paragraph{Comparing similarity measures}
As laid out in Section~\ref{ch2:methods:sim_probing}, we can construct task spaces based on different similarity measures (via performance transfers i.e. `\textit{transfer probing}' or properties of their gradients i.e. `\textit{gradient probing}').
Additionally, we consider multiple options to construct gradient task spaces:
the overlap between gradient subspaces $J_{sim}$ (i.e. the degree to which tasks are learned with the same parameters), the alignment of the gradients in those subspaces (using cosine similarity [$CS$]) or the gradient alignment weighted by the degree of subspace overlap (${J_{sim} \times CS}$). 
In the upper part of Table~\ref{tab:correlations_spaces}, we compare how predictable the resulting gradient task spaces (GTS) are for the respective TTS. 
Subspace overlap alone (GTS$_{J_{sim}}$) yields comparatively low correlations, as sharing parameters is necessary but not sufficient for transfer to occur. 
Accordingly, we find the similarity of gradients within the overlapping subspaces (GTS$_{CS}$) to be much more predictive of transfers between tasks.
If we now weigh the alignment by the degree of overlap (GTS$_{J_{sim} \times CS}$), we expect the GTS to become even more predictive of transfers (i.e. high alignment with larger parameter sharing should lead to higher transfer than high alignment with little parameter sharing).
Surprisingly, GTS$_{J_{sim} \times CS}$ does not lead to improved prediction of transfers leaving gradient alignment as the best predictor of transfers.

\begin{table}[h]
\setlength{\tabcolsep}{1.5pt}
    \centering
    \begin{tabular}{llccc}
        \toprule
        \rotatebox[origin=c]{\tablerotation}{\textbf{Task space}}& \rotatebox[origin=c]{\tablerotation}{\textbf{Hypothesis }} & \rotatebox[origin=c]{\tablerotation}{\normalsize{\textsf{LM27}}} & \rotatebox[origin=c]{\tablerotation}{\normalsize{\textsf{LM70}}} & \rotatebox[origin=c]{\tablerotation}{\normalsize{\textsf{LM203}}} \\
        \midrule 

        \multirow{3}{*}{TTS} & GTS$_{J_{sim}}$ & \normalsize{.41}\scriptsize{$\pm$.01} & \normalsize{.41}\scriptsize{$\pm$.01} & \normalsize{.39}\scriptsize{$\pm$.01} \\
        & GTS$_{CS}$ & \normalsize{.70}\scriptsize{$\pm$.02} & \normalsize{.72}\scriptsize{$\pm$.01} & \normalsize{.69}\scriptsize{$\pm$.02} \\
        & GTS$_{J_{sim} \times CS}$ & \normalsize{.51}\scriptsize{$\pm$.01} & \normalsize{.50}\scriptsize{$\pm$.01} & \normalsize{.46}\scriptsize{$\pm$.01} \\
        \midrule[0.5pt]
        \midrule 
        
        \multirow{3}{*}{TTS} & V. overlap  & \normalsize{.17}\scriptsize{$\pm$.03} & \normalsize{.15}\scriptsize{$\pm$.03} & \normalsize{.16}\scriptsize{$\pm$.01} \\
        & WD\tablefootnote{For WD, we report the absolute value of the correlation, as we relate a distance with a similarity measure.} & \normalsize{.13}\scriptsize{$\pm$.02} & \normalsize{.17}\scriptsize{$\pm$.01} & \normalsize{.13}\scriptsize{$\pm$.01} \\
        & by phen. & \normalsize{.27}\scriptsize{$\pm$.01} & \normalsize{.29}\scriptsize{$\pm$.01} & \normalsize{.33}\scriptsize{$\pm$.02} \\ 
        \midrule[0.5pt]
        \midrule 
        
        \multirow{3}{*}{\shortstack{GTS$_{CS}$}} & V. overlap  & \normalsize{.20}\scriptsize{$\pm$.01} & \normalsize{.19}\scriptsize{$\pm$.01} & \normalsize{.18}\scriptsize{$\pm$.01} \\
        & WD & \normalsize{.20}\scriptsize{$\pm$.01} & \normalsize{.25}\scriptsize{$\pm$.00} & \normalsize{.27}\scriptsize{$\pm$.01} \\
        & by phen. & \normalsize{.40}\scriptsize{$\pm$.00} & \normalsize{.43}\scriptsize{$\pm$.00} & \normalsize{.44}\scriptsize{$\pm$.00} \\
        \bottomrule
    \end{tabular}
    \caption[]{Correlations between task spaces and different hypothesis spaces. The first set of rows shows the correlations of the transfer task space (TTS) with gradient task spaces (GTS). GTSs are based on various similarity metrics (Jaccard Similarity [$J_{sim}$]; cosine similarity [$CS$]; the product of $J_{sim}$ and $CS$ [${J_{sim} \times CS}$]). GTS$_{CS}$ is the most predictive of transfers between linguistic tasks.
    The second and third sets of rows show the correlations of TTS and GTS$_{CS}$ with low-level controls (the shared vocabulary of different tasks [V. overlap] and the Wasserstein distance between tasks [WD]) as well as with the clustering-by-phenomena hypothesis space.
    Generalisation within phenomena is stronger than across low-level controls.
    }
    \label{tab:correlations_spaces}
\end{table}

\begin{figure*}[h!]
\centering

\begin{subfigure}[t]{0.32\linewidth} 
\includegraphics[width=\linewidth]{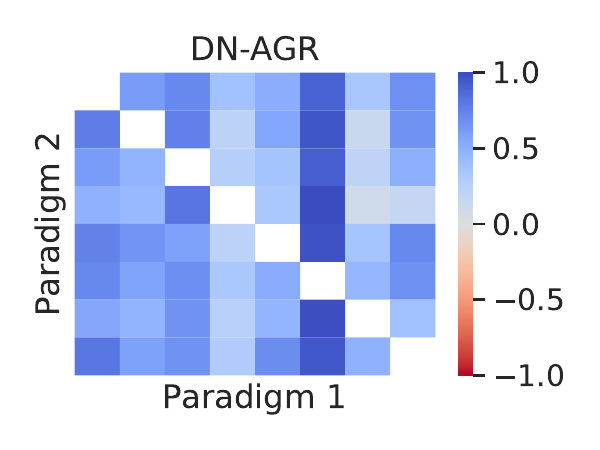}
\caption{}
\label{subfig:within_phenomenon_transfer_heatmaps_a}
\end{subfigure}
\hfill
\begin{subfigure}[t]{0.32\linewidth} 
\includegraphics[width=\linewidth]{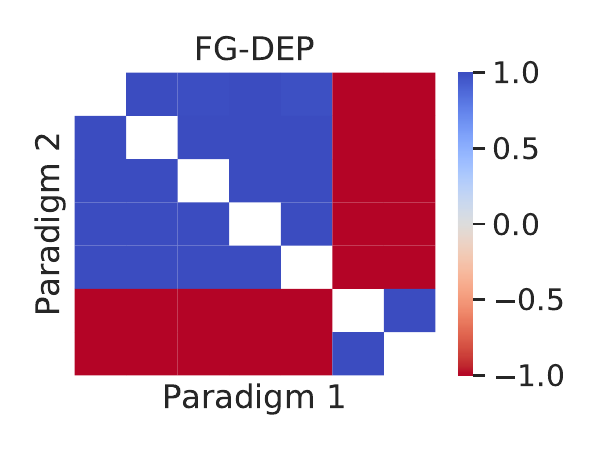}
\caption{}
\label{subfig:within_phenomenon_transfer_heatmaps_b}
\end{subfigure}
\hfill
\begin{subfigure}[t]{0.32\linewidth} 
\includegraphics[width=\linewidth]{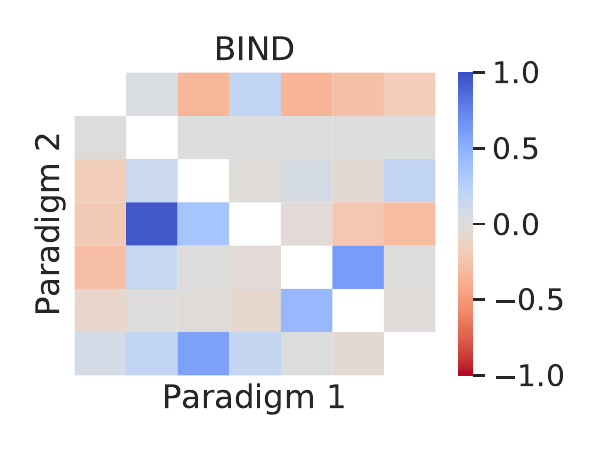}
\caption{}
\label{subfig:within_phenomenon_transfer_heatmaps_c}
\end{subfigure}
\caption{Different similarity patterns within phenomena for {\small\textsf{LM203}} after 20 epochs of pre-training. We find high similarity for all different paradigms in \phenomenon{determiner noun agreement} (a); high similarity but interfering subclusters for \phenomenon{filler-gap dependencies} (b); and no similarity for different \phenomenon{binding} paradigms (c). The exact identities of the individual rows and columns can be found in Table~\ref{tab:phenomena_paradigms} in Appendix~\ref{ch2:app:similarity_spaces}.}
\label{fig:within_phenomenon_transfer_heatmaps}
\end{figure*}

\begin{figure*}[h!]
\centering
\begin{subfigure}[t]{0.99\linewidth} 
\includegraphics[width=\linewidth]{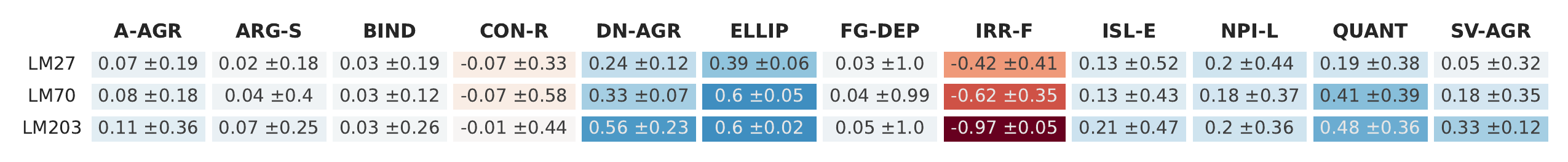}
\caption{}
\label{subfig:...}
\end{subfigure}

\caption{The degree of within-phenomena transfer for different models pre-trained for 20 epochs. A high value indicates that the model strongly generalises the phenomenon. A mapping of abbreviations to full names of phenomena can be found in Appendix~\ref{app:mapping_abbr_full_name}}
\label{fig:within_phenomenon_transfer_table}
\end{figure*}

\paragraph{Global transfer patterns}
 

We can get a global idea of the type of features the LMs generalise across, by comparing the task space with a `hypothesis space', a synthetic space representing a hypothesis that we \textit{expect} a model to generalise across.
To see whether the task spaces capture meaningful higher-level features within linguistic phenomena, or rather are due to low-level, spurious features, we generate three different types of hypothesis spaces.
First, we test the clustering of paradigms into their higher-level BLiMP phenomena as it is shown in Figure~\ref{fig:transfer_spaces}. 
A high correlation of task spaces with this hypothesis space means that LMs indeed find the higher-level structural features that all paradigms from the same phenomenon have in common.
We compare this against low-level, spurious controls in the form of vocabulary-level similarities between tasks (normalised vocabulary overlap [V. overlap] and the Wasserstein distance [WD] between vocabulary distributions; additional information in Appendix~\ref{ch2:app:control:vocab_baselines}).
We observe that the phenomenon structure is much more predictive of the generalisation patterns than the low-level controls (see the bottom portion of Table~\ref{tab:correlations_spaces}), confirming that the models generalise related tasks beyond their shared vocabulary and that we can capture this generalisation in our task spaces. 
The generalisation within phenomena further increases with increasing model size as shown by higher correlations for larger models.

\paragraph{Individual phenomena}
In the previous paragraph, we found that LMs tend to generalise according to the higher-level linguistic structure globally.
We can get a more differentiated picture by looking at the within-phenomena transfers for individual phenomena.
As laid out in §~\ref{ch2:methods:contr_tuning}, there are three main transfer patterns that we can expect within a phenomenon: 
First, the paradigms within a phenomenon have high similarity values, as we can see in the phenomenon \phenomenon{determiner noun agreement} [\textbf{DN-AGR}]) in Figure~\ref{subfig:within_phenomenon_transfer_heatmaps_a}). In this case, the model has discovered the overarching phenomena.
Beyond \textbf{DN-AGR}, we observe a similar pattern for \textbf{ELLIP}, \textbf{QUANT} and \textbf{SV-AGR}, showing in the high average within-phenomenon similarity values with relatively low standard deviations in Figure~\ref{fig:within_phenomenon_transfer_table} (see Appendix~\ref{app:mapping_abbr_full_name} for a full table of abbreviations).
Second, the model discovers the similarity between paradigms, but cannot reconcile them, as we see it in \phenomenon{filler-gap dependencies} [\textbf{FG-DEP}] (see Figure~\ref{subfig:within_phenomenon_transfer_heatmaps_b}). 
Those phenomena have low similarity values but very high standard deviations or negative similarity.
Models find subclusters of paradigms to transfer across, but cannot reconcile the different subclusters with each other. 
As a consequence, the subclusters are highly interfering with each other.
In irregular forms [\textbf{IRR-F}], the `subclusters' consist of single paradigms that test different usages of lexical items with irregular morphology (e.g. as a verb vs. as an adjective). Our models do not resolve this ambiguity, leading to high interferences between the tasks.
Third, for some phenomena, we do not observe any interactions between their paradigms (see, e.g. \phenomenon{binding} [\textbf{BIND}] in Figure~\ref{subfig:within_phenomenon_transfer_heatmaps_c}).
The LM finds idiosyncratic solutions to all the paradigms and does not discover the more general phenomenon. 
With increasing size, models tend more towards the first pattern, solving paradigms more by generalising to the higher-level phenomenon.

\begin{figure*}[!t]
  \begin{tabular}[b]{cc}
  \raisebox{5pt}{
    \begin{tabular}[b]{c}
      \begin{subfigure}[b]{0.22\linewidth}
        \includegraphics[width=\linewidth]{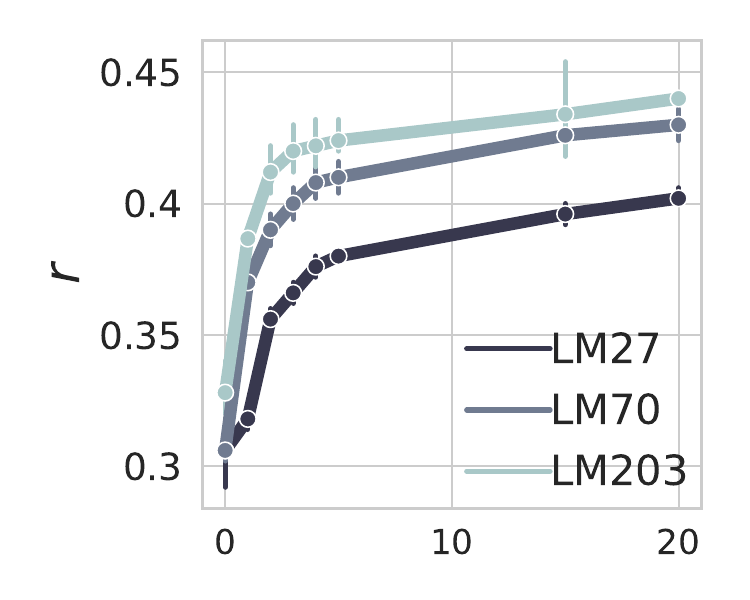}
        \caption{GTS vs. by phen.} \label{subfig:corr_phenomena_sim_space_throughout_training}
      \end{subfigure}\\
      \begin{subfigure}[b]{0.22\linewidth}
        \includegraphics[width=\linewidth]{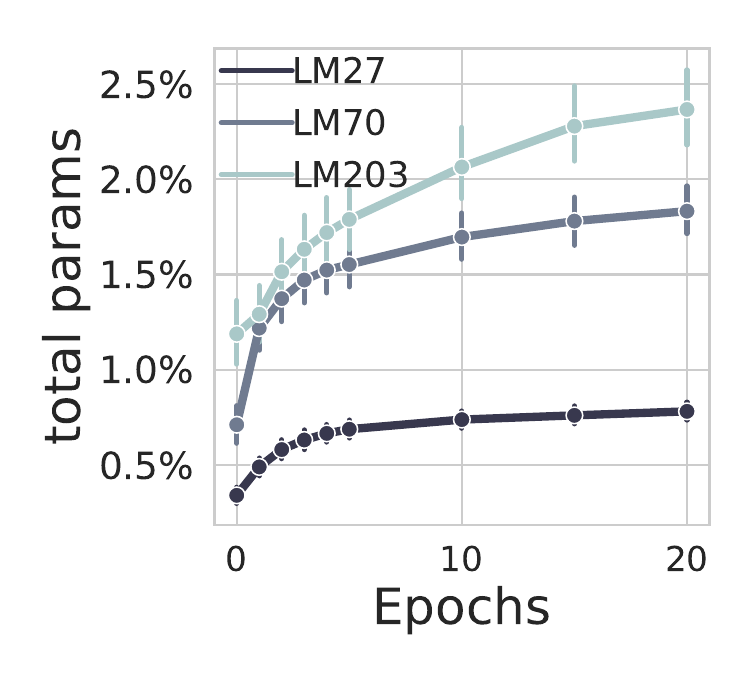}
        \caption{Average subspace size}
        \label{subfig:subspace_size_through_training}
      \end{subfigure}
    \end{tabular}}
    \hspace{-1cm}
    \hfill
    &
    \begin{subfigure}[b]{0.75\linewidth}
      \includegraphics[width=\linewidth]{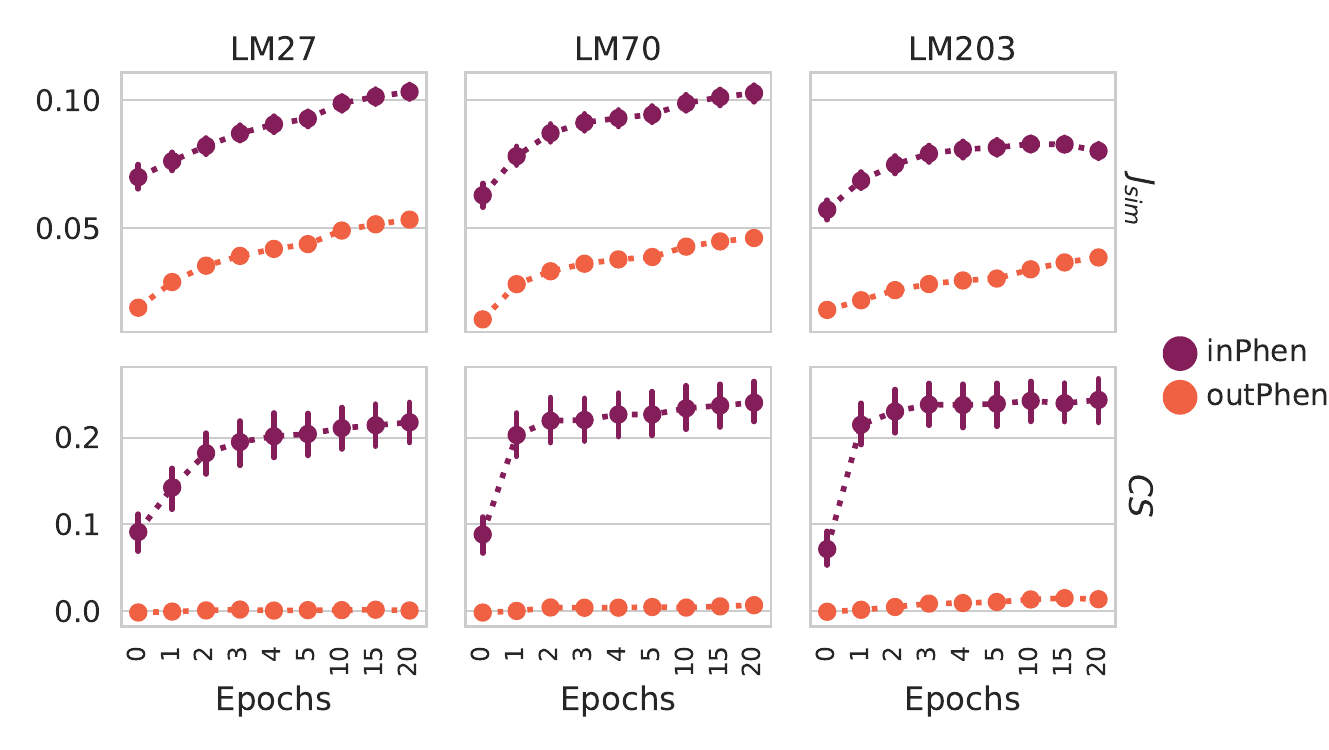}
      \caption{Average subspace overlap and gradient alignment}
      \label{subfig:cossim_and_jaccard_in_vs_out_phen}
    \end{subfigure}
  \end{tabular}
  \label{fig:ABC}
  \caption{(a) Correlation of GTS with clustered-by-phenomena space throughout training; (b) The development of subspace size throughout language model training. (c) (Top) The average $J_{sim}$ of task-subspaces either within the same phenomenon or outside the phenomenon; (bottom) the average inner product of $\Delta g$ of the overlapping subspaces. Larger models align related paradigms much faster and to a higher degree than smaller models.}
\end{figure*}

\subsubsection{Analysis of the training process}
\label{ch2:experiments:results:learning_process}
We have established that linguistic task spaces inform us about LMs language conceptualisation. 
But how do they change throughout training?
Performance on the BLiMP benchmark increases starkly in early training (for learning curves, see Appendix~\ref{ch2:app:BLiMP_learning_performance:BLiMP_learning_curves}).
What can we learn from task spaces beyond that observation?

\paragraph{Development of task spaces}
We construct similarity spaces for nine model checkpoints throughout pre-training (see Appendix~\ref{ch2:app:similarity_spaces} for visualisations of all similarity spaces).
Overall, we find that similarity spaces are remarkably stable: similarity patterns are present from very early in training (within the first epochs; for details, see Appendix~\ref{ch2:app:similarity_spaces_stability}).
At the same time, the generalisation within phenomena continuously increases, showing a continuous reinforcement of the existing generalisation pattern without major structure shifts in the pattern (see Figure~\ref{subfig:corr_phenomena_sim_space_throughout_training}). 

The continuity of generalisation patterns is surprising and contrasts with human learning, which is marked by learning stages \citep{piaget1952origins,gopnik1999scientist,gopnik2004theory}:
as humans deepen their knowledge, new patterns emerge. Language learning in LMs is not marked by such incisive shifts.

\paragraph{Development of subspaces}
Another interesting angle of analysis is the change of the subspaces $\theta_0$ in which the model is learning specific paradigms.
The average size of $|\theta_0|$ continuously grows during pre-training (see Figure~\ref{subfig:subspace_size_through_training}), showing that LMs learn linguistic tasks initially more localised but become more distributed throughout training. 
While the relative subspace overlap $J_{sim}$ increases generally with training, the within-phenomenon overlap is overall higher (see Figure~\ref{subfig:cossim_and_jaccard_in_vs_out_phen} top). 
Additionally, the alignment of gradients increases selectively for paradigms from the \textit{same} phenomenon (see Figure~\ref{subfig:cossim_and_jaccard_in_vs_out_phen} bottom).
This shows how the processing of linguistic tasks starts idiosyncratic (separated and in non-aligned subspaces) and with training, the sharing of structure increases (shared and in aligned subspaces, where appropriate).

\subsection{Linguistic hypothesis testing}
\label{ch2:experiments:linguistic_hypothesis_testing}
In addition to testing linguistic task spaces against hypothesis spaces with known structural similarities (such as `clustering by phenomena' or our vocabulary controls), we can also use them to test \textit{assumed} similarities in linguistic structure. 
We can construct a hypothesis space that represents contested ideas in linguistic theory and test whether our LM generalises according to the hypothesis by calculating a simple correlation. 
Our methodology is a step towards model-based theorising or `synthetic linguistics' \citep{chowdhury-zamparelli-2019-lstm}.
Doing linguistic hypothesis testing, however, is beyond the scope of the current paper.

\section{Discussion}
\label{ch2:discussion}

In this paper, we construct linguistic task spaces, representing an LM's language conceptualization, which can be used for linguistic hypothesis testing and as a holistic interpretability tool.
We introduce FTGD to selectively fine-tune latent, entangled concepts such as linguistic tasks, and `similarity probing' to estimate similarities between linguistic tasks through their transfer learning and gradient analysis.
We analyse the resulting similarity spaces of LMs throughout pre-training to interpret their learning process.
 
We find that learning of linguistic tasks begins localized and becomes more distributed with training, with increased parameter sharing among linguistic tasks and gradient alignment especially between linguistically related tasks.
Learning theory suggests that a trained model requires fewer dimensions due to increasingly efficient compression rules \citep[\textit{e.g.}][]{cheng2023bridging}. Opposing the assumed reduction in \textit{intrinsic} dimension, training in our experiments actually \textit{increases} the number of \textit{extrinsic} dimensions on which a task is learned.
Intrinsic and extrinsic dimensions might be inversely related in language models, as previously observed by \citet{aghajanyan2021intrinsic}.
A more distributed processing of tasks allows for more overarching structure sharing and generalisation across different subconcepts, which potentially helps to achieve lower \textit{intrinsic} dimensionality.
Furthermore, we find that generalization patterns remain surprisingly stable throughout pre-training, without stark shifts to new patterns—a behaviour more typical of human-like learning.
This potentially reflects the weakness of classical neural network models to generalise systematically \citep[][see also \citet{lake2018generalization,ettinger2018assessing,bahdanau2019closure,keysers2019measuring,yu2020assessing,kim2020cogs,press2022measuring}]{hupkes2020compositionality,lake2023human}.
Future generations of LMs employing more human-like learning paradigms \citep[see e.g.][]{lake2023human} may exhibit stronger shifts in generalization patterns.
The observed continuity might explain the lack of successful curriculum learning strategies for language modelling in the past \citep[see e.g.][]{surkov2021data, campos2021curriculum, weber2023curriculum}: in a learning process without notable shifts in generalisation patterns, changes in the data distribution during training are not beneficial.

\paragraph{Future reseach}
Beyond language, our approach to interpreting LM conceptualisation can be applied to other domains to better understand the current weaknesses of LMs, such as numerical reasoning and cross-lingual concept learning.
Furthermore, the potential for explicit linguistic hypothesis testing, though underexplored in this paper, can help bridge the gap between formal linguistic and computational linguistic research. Large, state-of-the-art LLMs may uncover subtle structural similarities that are informative to linguists.


\section{Limitations}
\label{ch2:limitations}
There are several limitations to the presented methods.
First, our fine-tuning and evaluation data are i.i.d. and come from a very narrow distribution: the data are not natural but synthetic, and all data are generated using the same templates. 
We use this very narrow i.i.d. data to asses the fine-tuning success during probing.
However, we cannot be entirely sure whether we succeeded in fine-tuning a specific linguistic task rather than some idiosyncracies of the narrow data distribution. 
While our FTGD approach might elevate this issue slightly, it does not dispel our doubts completely. The optimal way to guarantee our results would be the evaluation on a set from a separate distribution.

Second, while our approach applies to all types of knowledge domains, it requires \textit{minimal pairs} of tasks within that domain to fine-tune them selectively. 
Minimal pairs are primarily used in linguistics and are uncommon in other knowledge domains.

Third, as discussed in the previous section, a major weakness of our probing approach lies in the necessary top-down definition of `anchors' that we use to span the space.
We utilise human-defined tasks and relate them to each other. 
However, a more accurate linguistic space can probably be described by `anchors' that are defined through the model itself and span the conceptual space with maximal expressivity. 

\section*{Acknowledgements}
We thank the COLT group at UPF for the discussions and their useful feedback; Further, LW thanks the Department of Translation and Language Sciences at the University Pompeu Fabra for funding.
Further, this research was possible through computational resources acquired through a grant of the European Research Council (ERC) under the European Union’s Horizon 2020 research and innovation programme (grant agreement No. 101019291). This paper reflects the authors’ view only, and the ERC is not responsible for any use that may be made of the information it contains.
Ultimately, we thank the anonymous reviewers for their time and useful feedback!

\bibliography{references/anthology,references/ch1}

\newpage
\onecolumn

\appendix
\section{Appendix}
The supplementary material to this paper contains additional information about the control hypothesis spaces that we employ to verify the meaningfulness of our linguistic spaces (Appendix~\ref{ch2:app:controls}).
Further, we document the development of the LMs' performance on BLiMP in different scenarios (Appendix~\ref{ch2:app:BLiMP_learning_performance}).
Ultimately, we show all heatmaps for all transfer and gradient spaces for all models throughout the whole training process (Appendix~\ref{ch2:app:similarity_spaces}).

\subsection{Controls}
\label{ch2:app:controls}
We include control conditions and baselines for our experiments. 
This appendix section provides additional details.

\subsubsection{Vocabulary baselines}
\label{ch2:app:control:vocab_baselines}
We calculate two baselines to estimate the amount of transfer that is due to mere vocabulary overlap between different paradigms:
\begin{enumerate}
    \item \textit{Normalised vocabulary overlap (NVO)} between the vocabularies $V_A$ and $V_B$ of paradigms $A$ and $B$ -- calculated simply as the size of their intersection normalised by the maximum vocabulary overlap between any paradigms $X$ and $Y$: \[NVO = \frac{|V_A \cap V_B|}{\max(|V_X \cap V_Y|)}\]  
    \item \textit{Wasserstein distance}  \citep[WD;][]{kantorovich1960mathematical} between the vocabularies distributions.
These vocabulary controls can be correlated with any task space or hypothesis space. 
For example, the correlation between these controls and the transfer task spaces §~\ref{ch2:experiments:results:trans_space} indicates how much of the transfer between different paradigms can be attributed to the vocabulary overlap between tasks alone.
\end{enumerate}

\begin{figure*}[h!]
\centering
\includegraphics[width=0.90\linewidth]{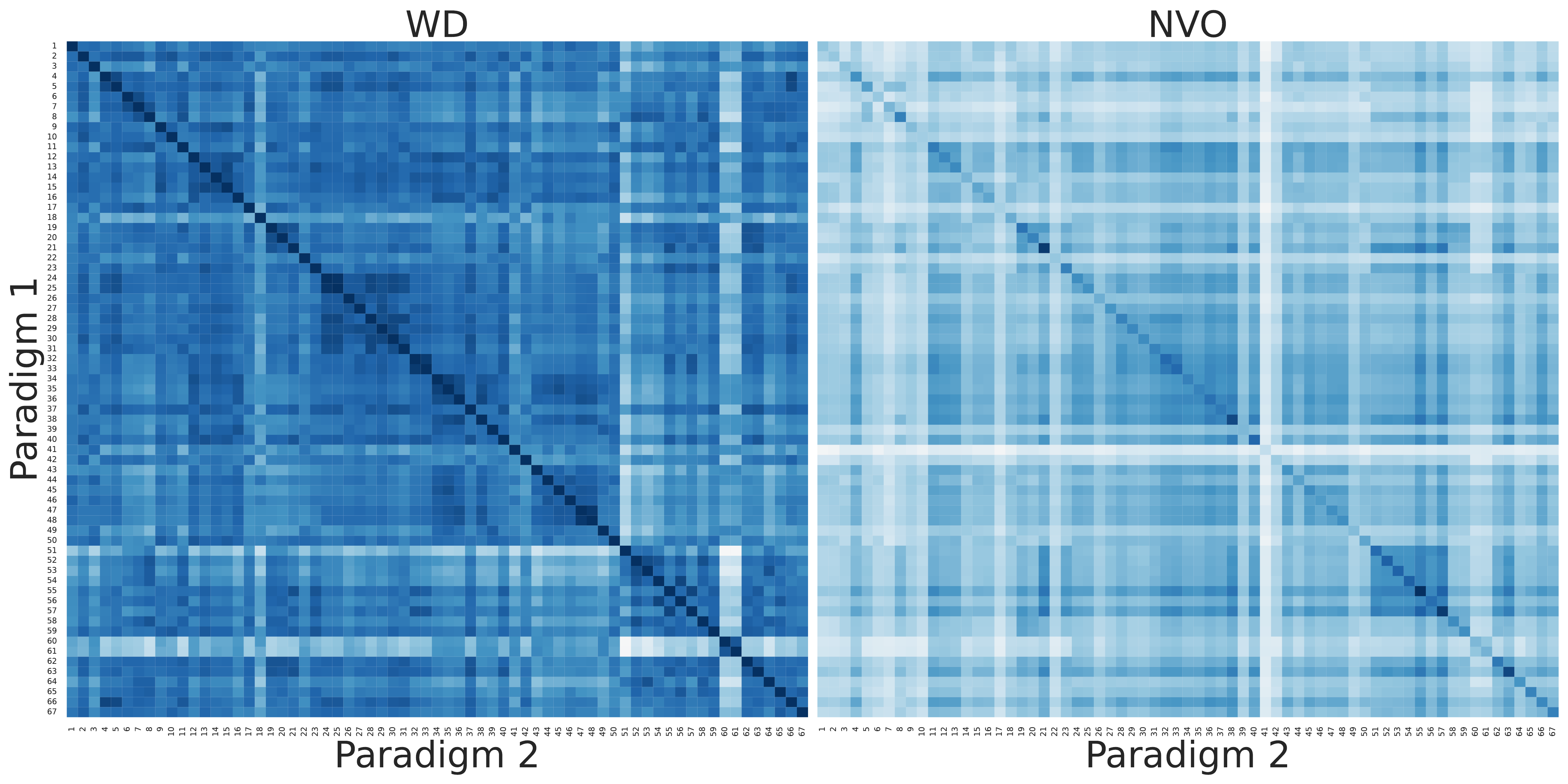}
\raisebox{10pt}{\includegraphics[width=0.08\linewidth]{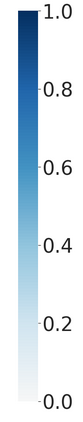}}
\caption{(1 - WD) in the left heatmap and the normalised vocabulary overlap on the right. Labels to individual rows and columns can be found in Table~\ref{tab:phenomenon_abbreviation}}
\label{fig:wasserstein_distance}
\end{figure*}

\subsection{BLiMP performance}
\label{ch2:app:BLiMP_learning_performance}
Throughout our experiments, we pre-train and fine-tune our LMs. 
We here document the performance of the models in different scenarios: first, we show how the models perform on the whole benchmark throughout the pre-training process.
Second, we show how different pre-training checkpoints adapt.

\subsubsection{BLiMP learning curves}
\label{ch2:app:BLiMP_learning_performance:BLiMP_learning_curves}
During the pre-training process, we evaluate each saved checkpoint on all paradigms of the BLiMP benchmark and average the results. The following plot shows the respective learning curves for the different models. 
While none of the models achieve very good performance, the largest model achieves their final performance much faster than the smaller ones.

\begin{figure*}[h!]
\centering
\includegraphics[width=0.55\linewidth]{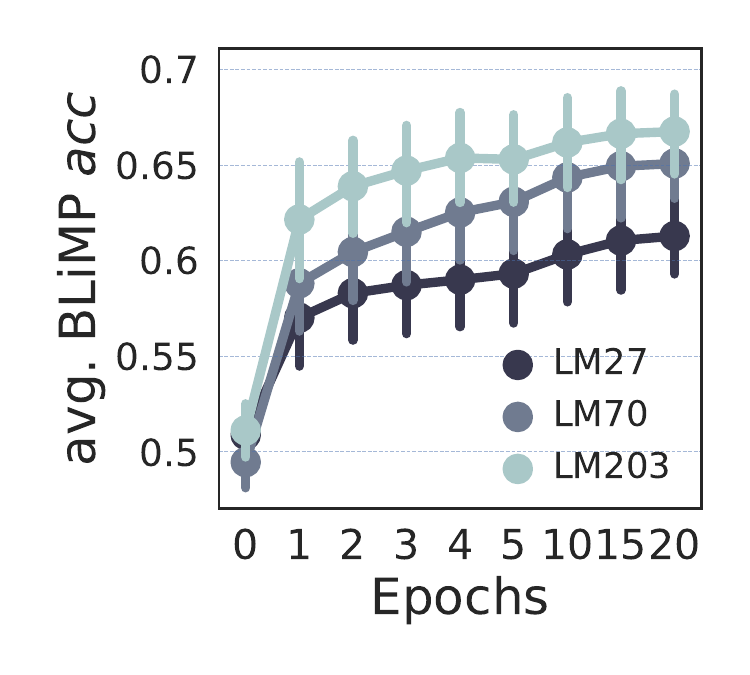}
\caption{Learning curves of our LMs on the BLiMP benchmark averaged across all paradigms and seeds.}
\label{fig:BLiMP_learning_curves}
\end{figure*}

\FloatBarrier
\subsubsection{BLiMP probe tuning}
\label{ch2:app:BLiMP_learning_performance:BLiMP_probe_tuning}
\begin{figure*}[h!]
\centering
\includegraphics[width=0.55\linewidth]{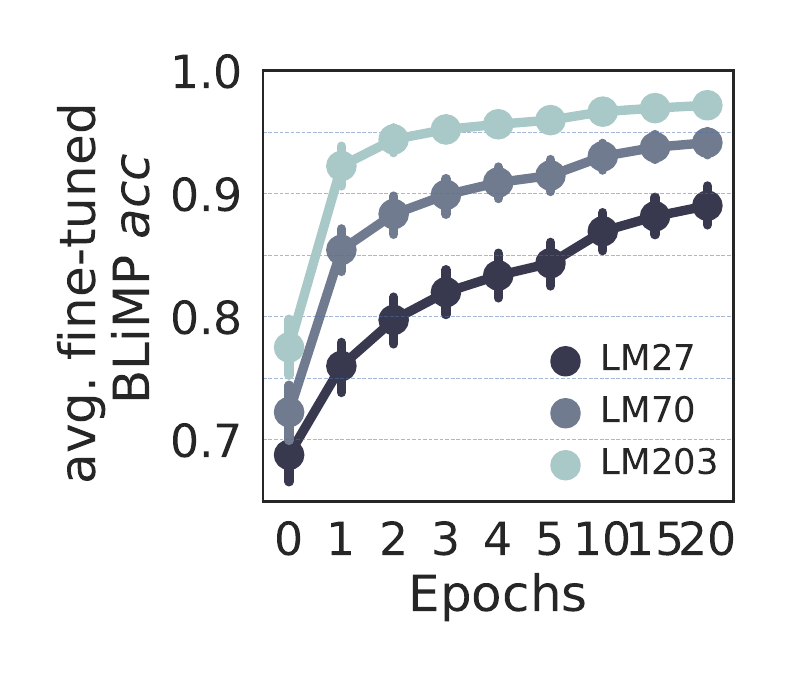}
\caption{Average final performance after FTGD a linguistic task.}
\label{fig:final_performance_finetuned}
\end{figure*}

The final performance after fine-tuning a specific task changes with the amount of pre-training.
The final performance of that model for that specific task is shown in Figure~\ref{fig:final_performance_finetuned}.
With more pre-training, models adapt better during the fine-tuning.
Larger models generally adapt better than smaller models.
FTGD works better for models that are pre-trained for longer.
This makes sense, as the method requires the difference between minimal pairs to be meaningful (i.e. it requires previous knowledge already contained in the model parameters). The subspace selection will be more accurate as a consequence.

\subsection{BLiMP abbreviation map}
\label{app:mapping_abbr_full_name}

\begin{table*}[h]
    \centering
    \begin{tabular}{llll}
        \toprule
        \textbf{Abbreviation} & \textbf{Phenomenon} & \textbf{Abbreviation} & \textbf{Phenomenon} \\
        \midrule
        A-AGR & Anaphor Agreement & ARG-S & Argument Structure \\
        BIND & Binding & CON-R & Control Raising \\
        DN-AGR & Determiner Noun Agreement & ELLIP & Ellipsis \\
        FG-DEP & Filler Gap Dependency & IRR-F & Irregular Forms \\
        ISL-E & Island Effects & NPI-L & NPI Licensing \\
        QUANT & Quantifiers & SV-AGR & Subject Verb Agreement \\
        \bottomrule
    \end{tabular}
    \caption[]{Mapping of abbreviations to linguistic phenomena.}
    \label{tab:phenomenon_abbreviation}
\end{table*}

\FloatBarrier
\subsection{Similarity space stability}
\label{ch2:app:similarity_spaces_stability}
We here show the correlation of similarity spaces of different epochs with the final similarity space (epoch 20).
Generally, similarity spaces are remarkably stable and do not show larger shifts in generalisation patterns.

\begin{figure}[h!]
\centering
\includegraphics[width=0.5\linewidth]{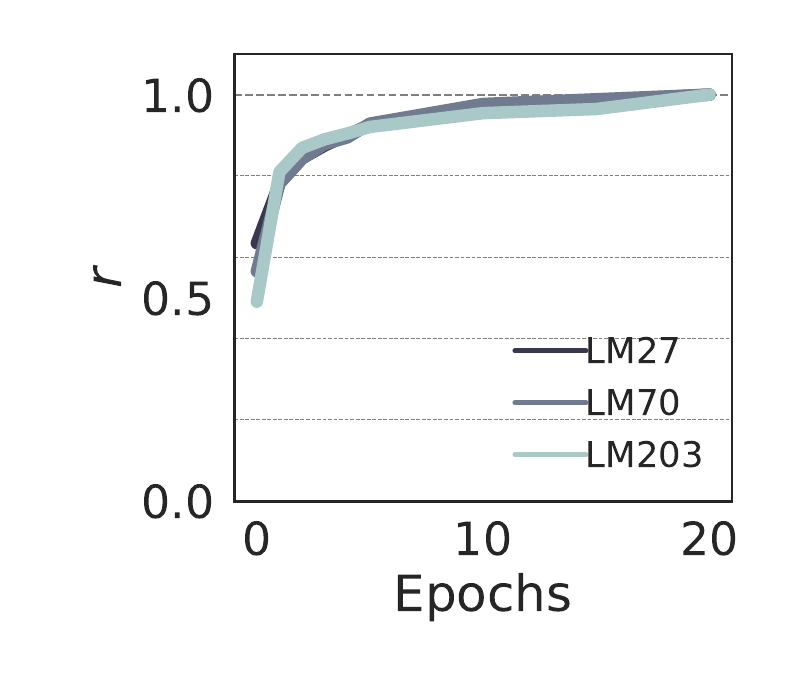}
\caption[TTS consistency throughout training]{Correlation of gradient similarity spaces with the trained-out gradient space. Correlation is very high after only a few epochs, indicating that the overall pattern of gradient similarities only changes minimally.}
\label{fig:consistency_cossim_space}
\end{figure}

\subsection{Details similarity spaces}
\label{ch2:app:similarity_spaces}
This section contains additional information about similarity spaces that we constructed throughout the paper.
\subsubsection{Similarity spaces through training}
\label{ch2:app:similarity_spaces_heatmaps}
\begin{sidewaysfigure}   

    \begin{subfigure}[b]{0.85\linewidth}
    \includegraphics[width=\linewidth]{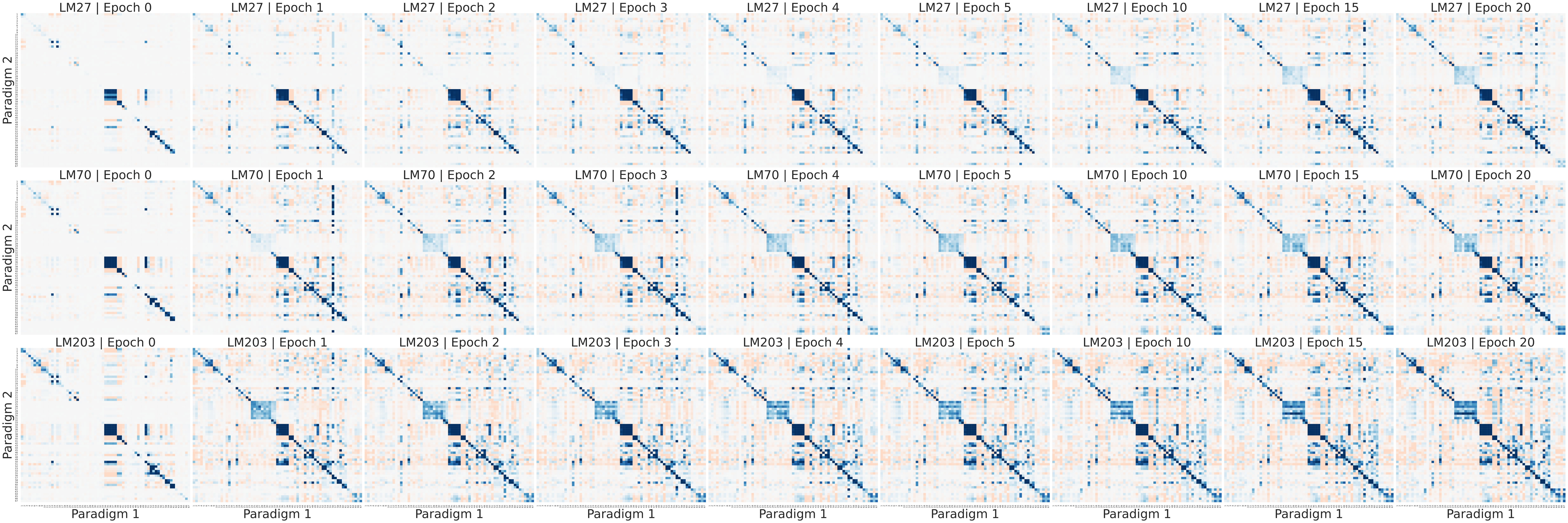}
    \caption{TTS for all models throughout training.} 
  \end{subfigure}
\begin{subfigure}[b]{0.85\linewidth}
    \includegraphics[width=\linewidth]{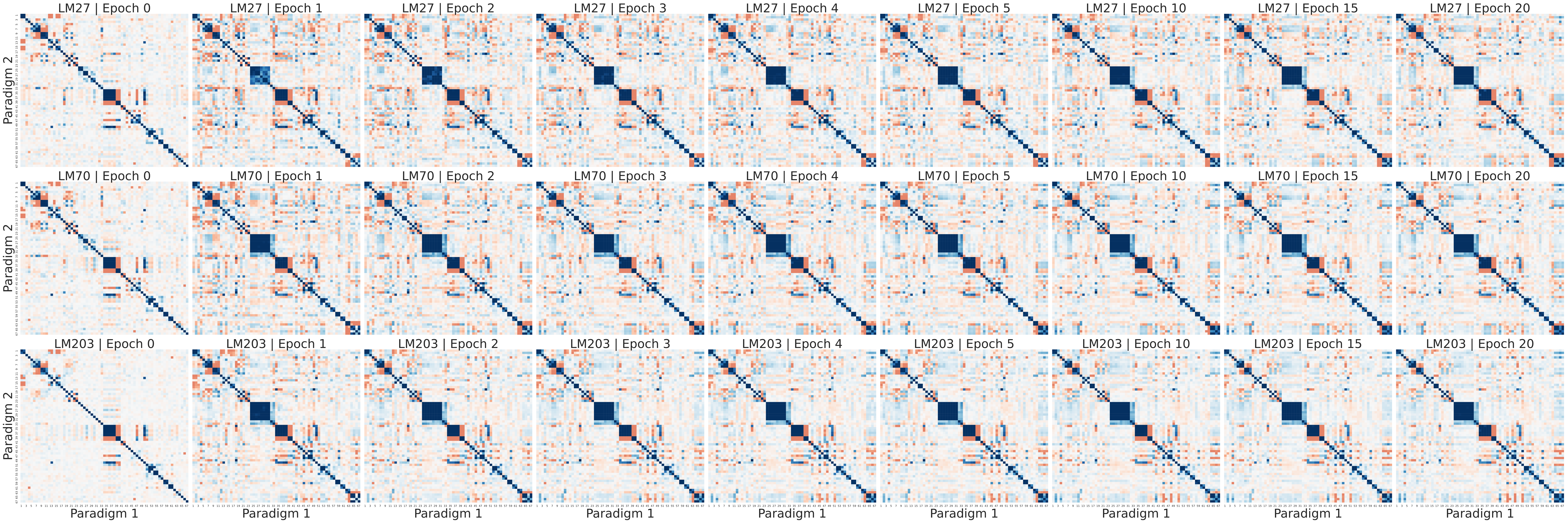}
    \caption{GTS for all models throughout training.} 
  \end{subfigure}%

    \caption[Transfer and gradient spaces for all checkpoints]{}
    \label{fig:transfer_spaces_all}   
\end{sidewaysfigure}

We construct all similarity spaces throughout the training process. 
Figure~\ref{fig:transfer_spaces_all} on the following page illustrates the transfer and gradient matrices for each saved model checkpoint.
The respective indices for the rows and columns of the heatmaps can be found in Table~\ref{tab:phenomena_paradigms} on the subsequent page.

\clearpage
\subsubsection{Paradigm/index map}
\label{ch2:app:paradigm_index_map}
A complete list of BLiMP phenomena, paradigms and the respective indices for the heatmaps throughout the paper.

\begin{table}[H]
\centering
\scriptsize
\begin{tabular}{lll}
\toprule
Phenomenon & Paradigm & Index \\
\midrule
\multirow{2}{*}{anaphor agreement} & anaphor gender agreement & 1 \\
 & anaphor number agreement & 2 \\
\cmidrule(lr){1-3}
\multirow{9}{*}{argument structure} & animate subject passive & 3 \\
 & animate subject trans & 4 \\
 & causative & 5 \\
 & drop argument & 6 \\
 & inchoative & 7 \\
 & intransitive & 8 \\
 & passive 1 & 9 \\
 & passive 2 & 10 \\
 & transitive & 11 \\
\cmidrule(lr){1-3}
\multirow{7}{*}{binding} & principle A c command & 12 \\
 & principle A case 1 & 13 \\
 & principle A case 2 & 14 \\
 & principle A domain 1 & 15 \\
 & principle A domain 2 & 16 \\
 & principle A domain 3 & 17 \\
 & principle A reconstruction & 18 \\
\cmidrule(lr){1-3}
\multirow{5}{*}{control raising} & existential there object raising & 19 \\
 & existential there subject raising & 20 \\
 & expletive it object raising & 21 \\
 & tough vs raising 1 & 22 \\
 & tough vs raising 2 & 23 \\
\cmidrule(lr){1-3}
\multirow{8}{*}{determiner noun agreement} & determiner noun agreement 1 & 24 \\
 & determiner noun agreement 2 & 25 \\
 & determiner noun agreement irregular 1 & 26 \\
 & determiner noun agreement irregular 2 & 27 \\
 & determiner noun agreement with adj 2 & 28 \\
 & determiner noun agreement with adj irregular 1 & 29 \\
 & determiner noun agreement with adj irregular 2 & 30 \\
 & determiner noun agreement with adjective 1 & 31 \\
\cmidrule(lr){1-3}
\multirow{2}{*}{ellipsis} & ellipsis n bar 1 & 32 \\
 & ellipsis n bar 2 & 33 \\
\cmidrule(lr){1-3}
\multirow{7}{*}{filler gap dependency} & wh questions object gap & 34 \\
 & wh questions subject gap & 35 \\
 & wh questions subject gap long distance & 36 \\
 & wh vs that no gap & 37 \\
 & wh vs that no gap long distance & 38 \\
 & wh vs that with gap & 39 \\
 & wh vs that with gap long distance & 40 \\
\cmidrule(lr){1-3}
\multirow{2}{*}{irregular forms} & irregular past participle adjectives & 41 \\
 & irregular past participle verbs & 42 \\
\cmidrule(lr){1-3}
\multirow{8}{*}{island effects} & adjunct island & 43 \\
 & complex NP island & 44 \\
 & coordinate structure constraint complex left branch & 45 \\
 & coordinate structure constraint object extraction & 46 \\
 & left branch island echo question & 47 \\
 & left branch island simple question & 48 \\
 & sentential subject island & 49 \\
 & wh island & 50 \\
\cmidrule(lr){1-3}
\multirow{7}{*}{NPI licensing} & matrix question NPI licensor present & 51 \\
 & NPI present 1 & 52 \\
 & NPI present 2 & 53 \\
 & only NPI licensor present & 54 \\
 & only NPI scope & 55 \\
 & sentential negation NPI licensor present & 56 \\
 & sentential negation NPI scope & 57 \\
\cmidrule(lr){1-3}
\multirow{4}{*}{quantifiers} & existential there quantifiers 1 & 58 \\
 & existential there quantifiers 2 & 59 \\
 & superlative quantifiers 1 & 60 \\
 & superlative quantifiers 2 & 61 \\
\cmidrule(lr){1-3}
\multirow{6}{*}{subject verb agreement} & distractor agreement relational noun & 62 \\
 & distractor agreement relative clause & 63 \\
 & irregular plural subject-verb agreement 1 & 64 \\
 & irregular plural subject-verb agreement 2 & 65 \\
 & regular plural subject-verb agreement 1 & 66 \\
 & regular plural subject-verb agreement 2 & 67 \\
\bottomrule
\end{tabular}
\caption{List of phenomena and paradigms}
\label{tab:phenomena_paradigms}
\end{table}

\end{document}